\documentclass[sigconf]{acmart}
\usepackage{booktabs}
\usepackage{arydshln}
\usepackage{caption}
\usepackage{multirow}
\usepackage{enumitem}

\AtBeginDocument{%
  }

\setcopyright{acmlicensed}
\copyrightyear{2018}
\acmYear{2018}
\acmDOI{XXXXXXX.XXXXXXX}
\acmConference[Conference acronym 'XX]{Make sure to enter the correct
  conference title from your rights confirmation email}{June 03--05,
  2018}{Woodstock, NY}
\acmISBN{978-1-4503-XXXX-X/2018/06}

\begin{document}


\title{Enhancing Temporal Sensitivity of Large Language Model for Recommendation with Counterfactual Tuning}

\author{Yutian Liu}
\email{liuyutian@mail.ustc.edu.cn}
\affiliation{%
  \institution{University of Science and Technology
of China}
  \country{China}
}

\author{Zhengyi Yang}
\email{yangzhy@mail.ustc.edu.cn}
\affiliation{%
  \institution{University of Science and Technology
of China}
  \country{China}}

\author{Jiancan Wu}
\email{wujcan@gmail.com}
\affiliation{%
  \institution{University of Science and Technology
of China}
  \country{China}
}

\author{Xiang Wang}
\email{xiangwang1223@gmail.com}
\affiliation{%
  \institution{University of Science and Technology
of China}
  \country{China}
}
\renewcommand{\shortauthors}{Trovato et al.}

\begin{abstract}

Recent advances have applied large language models (LLMs) to sequential recommendation, leveraging their pre‑training knowledge and reasoning capabilities to provide more personalized user experiences.
However, existing LLM‑based methods fail to sufficiently leverage the rich temporal information inherent in users' historical interaction sequences, stemming from fundamental architectural constraints: LLMs process information through self-attention mechanisms that lack inherent sequence ordering and rely on position embeddings designed primarily for natural language rather than user interaction sequences.
This limitation significantly impairs their ability to capture the evolution of user preferences over time and predict future interests accurately.

To address this critical gap, we propose \underline{C}ounterfactual \underline{E}nhanced \underline{T}emporal Framework for LLM‑Based \underline{Rec}ommendation (CETRec).
CETRec is grounded in causal inference principles, which allow it to isolate and measure the specific impact of temporal information on recommendation outcomes.
By conceptualizing temporal order as an independent causal factor distinct from item content, we can quantify its unique contribution through counterfactual reasoning—comparing what recommendations would be made with and without temporal information while keeping all other factors constant.
This causal framing enables CETRec to design a novel counterfactual tuning objective that directly optimizes the model's temporal sensitivity, teaching LLMs to recognize both absolute timestamps and relative ordering patterns in user histories.
Combined with our counterfactual tuning task derived from causal analysis, CETRec effectively enhances LLMs' awareness of both absolute order (how recently items were interacted with) and relative order (the sequential relationships between items).
Extensive experiments on real-world datasets demonstrate the effectiveness of our CETRec. Our code is available at https://anonymous.4open.science/r/CETRec-B9CE/.

\end{abstract}


  

\maketitle

\section{Introduction}



{In the era of information overload, recommender systems have become essential tools that help users navigate vast amounts of online content \cite{rec_survey_1}. Sequential recommendation, which models users' temporally ordered interaction histories to capture evolving preferences, has received increasing attention for producing more personalized and accurate suggestions \cite{rec_survey,Caser,GRU4Rec,SASRec,tiger}. More recently, the rapid advancement of large language models (LLMs) has motivated many works to integrate these models into sequential recommendation systems, leveraging their rich pre-trained knowledge and reasoning capabilities \cite{bigrec,LC-Rec,llara,P5,TALLRec,LLMRec}. Temporal information — the chronological order and timing of user interactions — remains central to sequential recommendation, since it directly reflects the evolution of user interests and preference dynamics.

}

\begin{figure}[t]
  \centering
  \includegraphics[width=\linewidth]{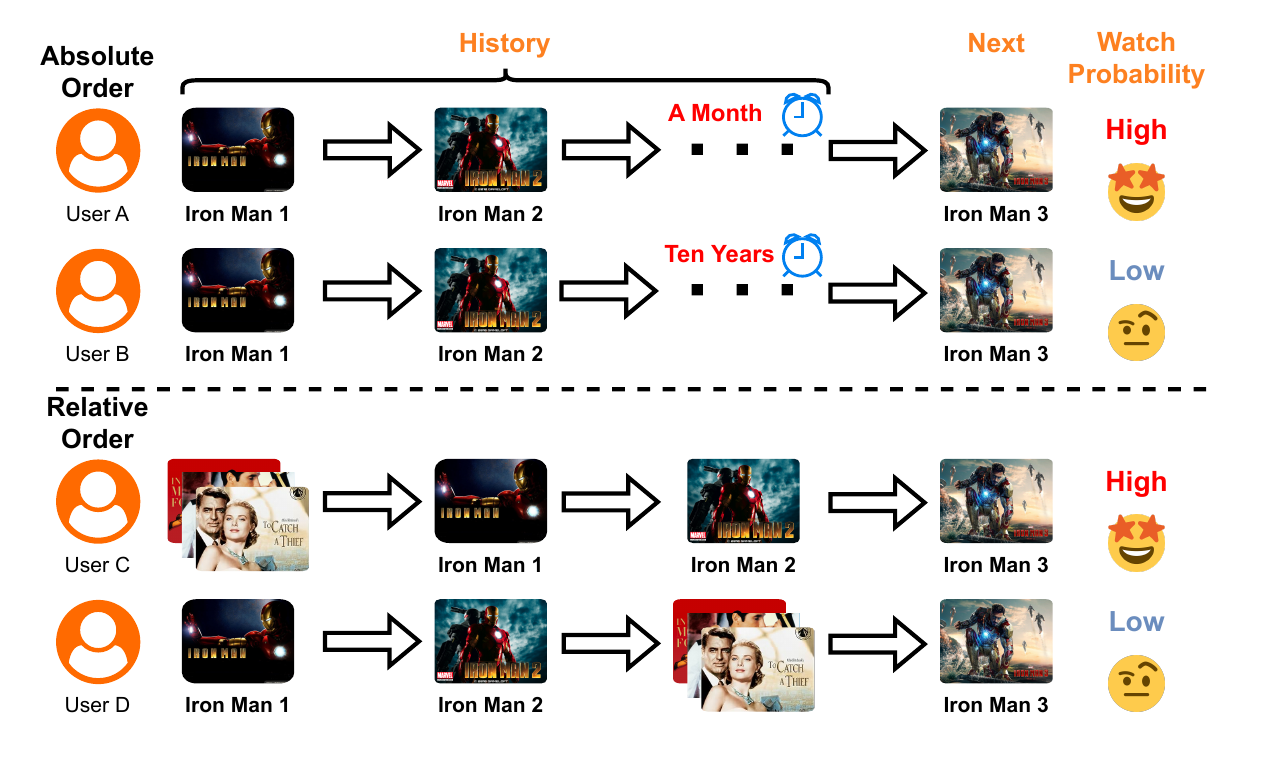}
  \vspace{-10pt}
  \caption{Illustration of absolute vs. relative order effects. Users A and B have the same history movie set but differ in absolute order; Users C and D have the same history movie set but differ in relative order.}
  \vspace{-10pt}
  \label{fig:1}    
  \Description{Different user interest shown in different interaction sequences.}
\end{figure}

In sequential recommendation, user historical interactions encompass two aspects of temporal information: \textbf{absolute order information} and \textbf{relative order information}. 
Absolute order refers to the actual timestamps of interactions (\textit{i.e.}, how long ago each event occurred), whereas relative order refers to the ordinal relationships among users' multiple interactions. 
Both aspects play a crucial role in predicting the next item a user may interact with. 
For example, as shown in Figure~\ref{fig:1}, user A who watched \textit{Iron Man} and \textit{Iron Man 2} within the last month is significantly more likely to proceed to \textit{Iron Man 3} than user B whose viewing occurred ten years ago. Moreover, if user D watches several romance films immediately after \textit{Iron Man} and \textit{Iron Man 2}, their probability of watching \textit{Iron Man 3} is typically lower than that of user C who views those same romance films before watching the first two \textit{Iron Man} installments, because the former’s interests may already have shifted.


{
Existing LLM-based recommendation approaches can be largely grouped into two main streams: 1) Paradigm adaptation methods reformulates recommendation as natural language generation (e.g., text completion or response generation) using prompting or fine-tuning to align LLMs’ knowledge with recommendation goals \cite{bigrec,P5}.
2) Multimodal integration represents items as a distinct modality and aligning these embeddings with LLMs’ text representations via cross-modal attention or embedding alignment \cite{recinterpreter,collm}.
Despite these advancements, these approaches primarily focus on modeling the semantic aspects of interacted items or optimizing the overall recommendation task, and do not explicitly capture the temporal structure of user interaction sequences. This temporal oversight leaves a significant gap in their ability to model users' evolving interests and time-dependent preference shifts.Several studies have exposed this weakness: TIN \cite{tin} analyzed the correlation between item position and recommendation accuracy, LLM-SRec \cite{lostinseq} perturbed interaction orders and observed minor performance drops compared to traditional models, these findings collectively suggest that LLM-based recommenders have limited temporal awareness despite diverse modeling strategies.
}


{Large language models, built on Transformer architectures, fundamentally rely on the self-attention mechanism which inherently lacks any built-in notion of sequence order \cite{Transformer}. 
While these models incorporate position embeddings as an external means of injecting token order for natural language processing tasks, such mechanisms prove inadequate for capturing the complex temporal dynamics in sequential recommendation.
This inadequacy stems from several factors: First, position embeddings in LLMs are designed primarily for natural language, where grammatical structure and semantic context provide additional ordering cues that are absent in item interaction sequences. 
Second, these embeddings are typically applied at the token level rather than the item level, preventing the model from treating each interacted item as a cohesive temporal unit during inference. 
Third, recommendation prompts typically include extensive task instructions and contextual background information, causing the already sparse temporal signals to become "diluted" amid multiple sources of information. 
As a result, the model struggles to correctly interpret the temporal relationships between user interactions, limiting its ability to capture the dynamic evolution of user preferences that is essential for accurate sequential recommendation. This fundamental limitation suggests that standard position embeddings in LLMs, while sufficient for natural language tasks, cannot adequately capture the nuanced temporal relationships crucial for effective sequential recommendation.
}


{
To address this challenge, we propose \underline{C}ounterfactual \underline{E}nhanced \underline{T}emporal Framework for LLM‑Based \underline{Rec}ommendation (CETRec), which leverages causal inference to explicitly model and enhance temporal sensitivity in recommendation. 
CETRec is grounded in a formal causal model that distinguishes temporal order as a distinct causal factor from the items themselves in determining recommendation outcomes.
This causal framing allows us to isolate and measure the specific impact of temporal information through counterfactual reasoning.
To model this temporal information, for each item, we assign every token the same additional position embedding on top of its original positional embedding, called temporal embedding.
Subsequently, by constructing a counterfactual world where all temporal order information is erased, specifically, replacing each item's temporal embedding with that of the first item, while keeping all other factors constant, we can precisely quantify the causal effect of temporal order on recommendations.
Based on this counterfactual inference, we design a novel tuning objective that maximizes the divergence between predictions in the factual (temporally ordered) and counterfactual (temporally erased) worlds. This approach directly optimizes the model's sensitivity to temporal patterns by encouraging it to produce different predictions when temporal information is altered, effectively teaching the model to recognize and utilize both absolute and relative temporal signals in user interaction sequences.
}

In summary, our contributions are threefold:
\begin{itemize}[leftmargin=*]
\item We reveal that existing LLM-based sequential recommenders insufficiently exploit temporal information, due to Transformer architecture limitations and position embeddings designed for NLP.
\item We propose a novel framework, \underline{C}ounterfactual \underline{E}nhanced \underline{T}empo-ral Framework for LLM‑Based \underline{Rec}ommendation (CETRec), which employs a principled causal approach to recommendation. CETRec integrates temporal embedding at the item level and introduces a counterfactual tuning task derived from causal inference that directly optimizes the model's sensitivity to both absolute and relative temporal ordering. By formalizing temporal order as an independent causal factor in our structural causal graph, we enable precise measurement and enhancement of its contribution to recommendation outcomes.
\item We conduct extensive experiments across multiple real-world datasets, validating the effectiveness of CETRec in enhancing temporal information comprehension. Through ablation analysis, we further verify CETRec's superior ability to leverage temporal patterns in generating recommendations that align with users' evolving interests.
\end{itemize}

\section{Related Work}

\subsection{Sequential Recommendation}

Sequential recommendation predicts the next item a user may be interested in by leveraging user and item features extracted from the user's historical interaction sequence. Early methods relied on Markov chains to capture transition probabilities between items \cite{Markov}. With the advent of deep learning, a variety of architectures---RNN \cite{GRU4Rec, RNN1, RNN2}, CNN \cite{Caser, CNN1, CNN2} and attention \cite{SASRec,attn1}---have been employed to model user–item interaction dynamics.

Recently, large language models (LLMs) have achieved remarkable success across domains \cite{llmsuervey,llmnlp,llmnlp1,llmrob,llmir,llmir1} thanks to two key strengths: their rich pre-training knowledge and powerful reasoning capabilities. Following these strengths, LLM‐based approaches to sequential recommendation broadly fall into two paradigms:
\begin{itemize}[leftmargin=*]
\item\textbf{Representation Enhancement}. These methods leverage the LLMs' pre-trained knowledge to improve item or user representations—either by extracting embeddings from items’ textual (or other modal) representations \cite{recinterpreter,repitem,repitem1,repitem2}, or by distilling embeddings from traditional models into the LLMs \cite{llara,LC-Rec,lostinseq}.

\item \textbf{Paradigm Adaption}. These methods leverage the reasoning capabilities of LLMs by aligning them with the recommendation paradigm—reformulated as natural language generation tasks—through techniques such as fine-tuning~\cite{P5,bigrec,TALLRec,gpt4rec}, prompting~\cite{prompt, tempura}, or in-context learning (ICL)~\cite{icl}.
\end{itemize}
Several recent studies have recognized the insufficient utilization of temporal information in interaction sequences by existing recommender systems.  
TIN~\cite{tin} examined the correlation between item positions in the sequence and the recommendation results, while LLM-SRec~\cite{lostinseq} evaluated model performance after shuffling the order of interacted items.  
Both studies revealed clear deficiencies in existing models.  
To address this issue, TIN adopts Target-aware Temporal Encoding (TTE), integrating it with item ID embeddings to capture semantic–temporal correlations.  
LLM-SRec distills item embeddings from traditional sequential recommenders, thereby injecting temporal information into LLMs.  
Tempura~\cite{tempura} designs a prompting framework to simulate human reasoning processes, aiming to enhance the temporal awareness of LLMs.

Building on these observations, CETRec injects unified item-level temporal embeddings directly at the LLM input, rather than distributing temporal signals across token-level embeddings or relying solely on distilled representations. 
This design preserves each item's lexical semantics while explicitly encoding both absolute and relative temporal relationships, thereby improving the model's temporal sensitivity and yielding more accurate, time-aware recommendations 


\subsection{Causal Inference for Recommendation}
Non-causal recommenders typically capture correlations only, which can produce spurious relationships and poor generalization; causal inference instead seeks the underlying data-generating mechanisms, yielding more robust and interpretable recommendations. Prior causal work has concentrated on three practical directions: debiasing (modeling popularity, position/click, and other exposure biases as confounders and applying interventions to correct them) \cite{debiased1,debiased,posbiat,cloickbiat}, using structural causal models to mitigate sparsity and noise by separating genuine preferences from random or missing interactions \cite{casualdata1,casualdata2,casualdata3}, and moving beyond accuracy toward explainability and fairness by identifying causal drivers of user behavior and detecting/mitigating discriminatory effects \cite{explain,explain1,fairness,fariness1}.

Unlike these prior efforts, our work brings causal inference into sequential recommendation with an explicit focus on temporal dynamics: we use counterfactual inference to isolate the causal effect of item order on recommendation outcomes, construct counterfactual worlds that erase temporal signals while preserving item content to measure temporal sensitivity, and introduce tuning objectives that improve LLMs’ modeling of both absolute and relative temporal patterns in user interaction sequences.
\section{Preliminaries}

In this section, we formally define our task and introduce the basic conceptions used in CETRec, including position embedding and causal model.
\begin{table}
  \centering
  \caption{A tuning template for the Game dataset. <His\_Seq> and <Target\_Item> denote the user's historical interaction sequence and the ground-truth next item, respectively.}
  \label{tab:prompt}

  \begin{tabular}{@{}p{0.3\linewidth}p{0.6\linewidth}@{}}
    \toprule
    \multicolumn{2}{c}{\textbf{Instruction Input}} \\
    \midrule
    \textbf{Task Instruction:} 
      & Given a list of video games the user has played before, please recommend a new video game that the user likes to the user. \\
    \addlinespace
    \cdashline{1-2}[2pt/2pt]
    \noalign{\vskip 4pt}\textbf{Task Input:} 
      & The user has played the following video games before: \texttt{<His\_Seq>} \\
    \midrule
    \multicolumn{2}{c}{\textbf{Instruction Output}} \\
    \midrule
    \textbf{Output:}
      & \texttt{<Target\_Item>} \\
    \bottomrule
  \end{tabular}
\end{table}
\subsection{Task Formulation}

In sequential recommendation scenario, the model predicts the next item a user is likely to interact with based on their historical interaction sequence.
Formally, let $\mathcal{I} =\{i_1,i_2,...,i_{|\mathcal{I}|}\}$ denote the set of items. Given a temporally ordered interaction sequence $\mathcal{S}= (i_{t_1},i_{t_2},...,i_{t_{|\mathcal{S}|}})$, where $i_{t_n} \in \mathcal{I}$ denotes the item interacted with at time $t_n$, the goal of the model is to predict the next item $i_{t_n+1}$.

\begin{figure*}[t]
    \centering
    \includegraphics[width=\linewidth]{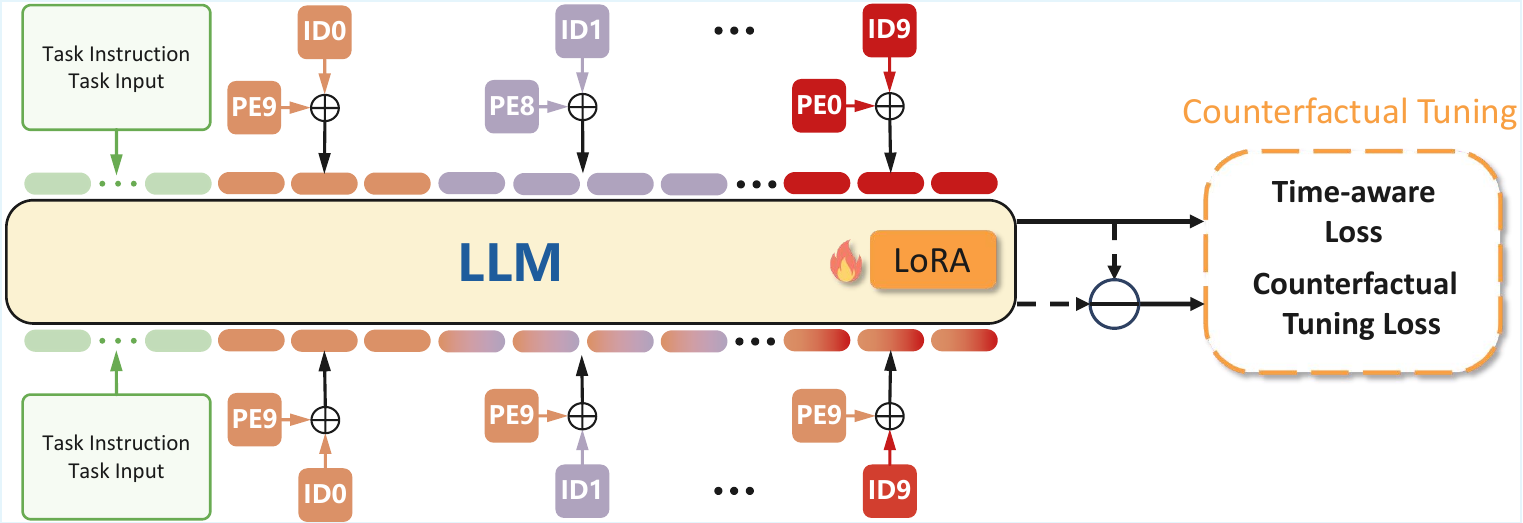}
    \vspace{-10pt}
    \caption{The framework of CETRec. The user interaction sequence is transformed to the instruction tuning format. CETRec constructs temporally ordered and erased sequence through assigning the corresponding position embedding to every token of each item. During counterfactual tuning, CETRec yield Time-aware Loss from temporal oredered sequence and counterfactual tuning loss from the difference between temporally ordered and erased sequence.}
    \vspace{-10pt}
    \label{fig:frame}
\end{figure*}

In this work, to enable the LLM to generate the target item’s title more accurately, we employ instruction tuning \cite{instruc}. The prompt template is illustrated in Table~\ref{tab:prompt}. In tuning phase, the user interaction sequence $\mathcal{S}$ replaces <His\_Seq>, forming the instruction input $x$, and the ground-truth next item consists the instruction output $y$. Notably, all items are represented by their title. Subsequently, the origin tuning loss$L_o$ can be computed as:
\begin{equation}
    \label{eq:1}
L_o=\sum_{(x,y)\in D}\sum_{k=1}^{|y|}\ell\left(f_\Theta(x, y_{<k});\ y_k\right),
\end{equation}
where $\ell$ denotes the loss function, $f_\Theta(\cdot)$ denotes the LLM with parameters $\Theta$, $y_k$ refers to the $k$-th token in $y$, and $y_{<k}$ represents tokens before $y_k$.

However, fine-tuning LLMs with full parameters is both time-consuming and computationally expensive. To address this, we adopt LoRA (Low-Rank Adaptation) \cite{lora}, which freezes the original model weights and trains only a small set of additional low-rank matrices, achieving comparable performance with a fraction of the overhead. Formally, The new loss is formulated as:
\begin{equation}
\label{eq:2}
L_n=\sum_{(x,y)\in D}\sum_{k=1}^{|y|}\ell\left(f_{\Theta+\Phi}(x, y_{<k});\ y_k\right),
\end{equation}
where $\Phi$ is the LoRA parameters.

\subsection{Position Embedding}
Position Embedding(PE) is a method used in Transformer-based models to encode the order of input tokens. Existing position embedding can be broadly categorized into two types: absolute and relative \cite{positionembedding}. In \textbf{absolute position embedding}, each input token embedding $x_k$ at position $k$ is augmented by a learnable or fixed vector $p_k$ that depends solely on $k$, typically of the form $x_k + p_k$. In \textbf{relative position embedding}, position information is incorporated dynamically within the attention mechanism by considering the relative distances between tokens, instead of adding fixed positional signals to the input embeddings.

\textbf{Sinusoidal Position Embedding(SinPE)} is a classic form of absolute position embedding proposed in the original Transformer architecture \cite{Transformer}, defined as:
\begin{equation}
    \label{eq:3}
\begin{cases} 
p_{k,2t}   &= \sin\left( \frac{k}{10000^{2t/d}} \right) \\ 
p_{k,2t+1} &= \cos\left( \frac{k}{10000^{2t/d} }\right) 
\end{cases},
    \end{equation}
where $p_{k,2t}$ is the $2t$-th element of the d-dimensional vector $p_k$.

\textbf{Rotary Position Embedding(RoPE) \cite{RoPE}} is a state-of-the-art relative position embedding method that incorporates characteristics of absolute position embedding. It injects positional information by rotating the query ($q$) and key ($k$) vectors during the attention computation based on their positions. The formulation is as follows:
\begin{equation}
    \label{eq:4}
f_{\{q,k\}}(x_{m}, m) = R_{\Theta, m}^{d} W_{\{q,k\}} x_{m},
\end{equation}
where $R_{\Theta, m}^{d}$ is A block-diagonal $d\times d$ matrix composed of $d/2$ independent $2\times 2$ rotation blocks, with $\Theta$ denoting rotary parameters associated with the position $m$ and each corresponding embedding dimension. Applying RoPE to self-attention process, we have:
\begin{equation}
    \label{eq:rope}
q_{m}^{\top}k_{n} = (R_{\Theta,m}^{d} W_{q} x_{m})^{\top} (R_{\Theta,n}^{d} W_{k} x_{n}) = x_{m}^{\top} W_{q} R_{\Theta, n-m}^{d} W_{k} x_{n},
    \end{equation}
where $R_{\Theta, n - m}^{d} = (R_{\Theta, m}^{d})^{\top} R_{\Theta, n}^{d}$ ,which only depends on the relative distances between tokens.
In this work, we employ the two position embedding schemes described above: SinPE and RoPE.

\subsection{Causal model.}
A causal model \cite{causality} formally describes the relationships between variables through a structured framework, typically defined as a triple:
$$M = \langle U, V, F \rangle, $$
where:
\begin{itemize}[leftmargin=*]
    \item \( U \) is the set of exogenous variables determined by factors outside the model;
    \item \( V \) is the set of endogenous variables determined within the model;
    \item \( F \) is the set of structural functions such that each \( f_i \in F \) maps \( U_i \cup \mathrm{PA}_i \) to \( V_i \), where \( U_i \subseteq U \), \( \mathrm{PA}_i \subseteq V \setminus V_i \), and the complete set \( F \) defines a mapping from \( U \) to \( V \).
\end{itemize}

Every causal model \( M \) is associated with a \textbf{causal graph} \( G(M) \), a directed graph where nodes represent variables, and edges are drawn from \( U_i \) and \( \mathrm{PA}_i \) to \( V_i \). 


\section{Methodology}
In this section, we present a comprehensive framework for enhancing temporal sensitivity in LLM-based sequential recommendation. We first conduct a causal analysis of sequential recommendation from a temporal perspective, establishing the theoretical foundation for our approach. Next, we introduce our novel method for modeling temporal order with position embeddings. We then apply counterfactual inference to quantify the causal effect of temporal information, which leads to the design of our CETRec framework, as illustrated in Figure~\ref{fig:frame}.

\begin{figure}[t]
  \centering
  \includegraphics[width=\linewidth]{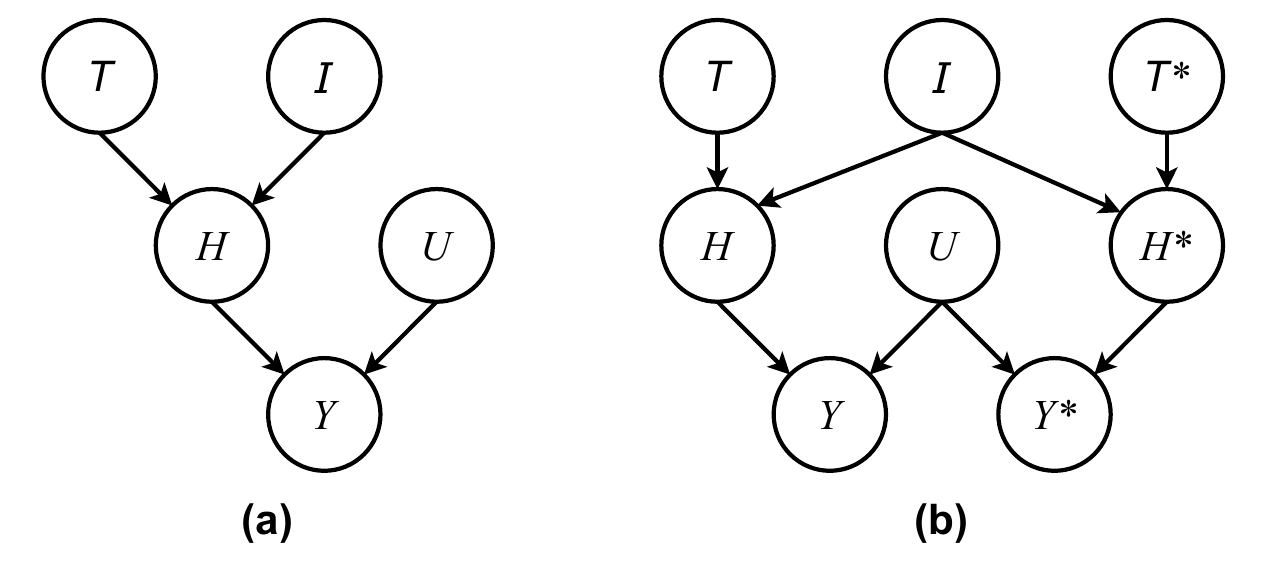}
  \vspace{-10pt}
  \caption{(a) Causal graph for sequential recommendation task. (b) The twin network of counterfactual inference on temporal order $T$.}
  \vspace{-10pt}
  \label{fig:Causal Anal}    
  \Description{Causal Anal}
\end{figure}
\subsection{Causal Analysis}
Sequential recommendation inherently involves complex temporal dynamics that influence user preferences. To systematically analyze these dynamics, we treat temporal order as an independent variable that, together with the interacted items themselves, defines the user's historical interaction sequence.
From this perspective, we perform a formal causal analysis of the sequential recommendation task and derive the causal graph shown in Figure~\ref{fig:Causal Anal}(a). This graph captures the key variables and their interdependencies in the recommendation process. The semantics of the nodes are as follows:
\begin{itemize}[leftmargin=*]
  \item \textbf{Node \(I\)}: items in the user's interaction history, representing the content aspects of historical interactions regardless of their temporal arrangement.
  \item \textbf{Node \(T\)}: temporal order of interacted items, encompassing both absolute timestamps and relative sequential relationships.
  \item \textbf{Node \(H\)}: the user's entire historical interaction sequence, which is jointly determined by both the items and their temporal ordering.
  \item \textbf{Node \(U\)}: background or contextual information affecting the recommendation results, such as shifts in user preference, changes in item popularity, or variations in item price/reputation over time.
  \item \textbf{Node \(Y\)}: he next item predicted or generated by the recommender system, which is the target variable we aim to optimize.
\end{itemize}
This causal graph provides several important insights. 
Previous studies have primarily focused on the path $H\rightarrow Y$ or path $I \rightarrow H\rightarrow Y$, corresponding to paradigm adaptation methods and multi-modal integration methods, respectively. 
The former treats recommendation as a sequence-to-item generation task, while the latter enriches item representations with specialized embeddings. 
However, the influence of temporal order---captured by the path $T \rightarrow H\rightarrow Y$---has been largely overlooked despite its crucial role in shaping user preferences and interaction patterns.

By explicitly modeling temporal order as a distinct causal factor, we can isolate its specific contribution to recommendation outcomes, allowing us to design targeted interventions that enhance LLMs' temporal sensitivity. This causal framing represents a significant departure from previous approaches that implicitly blend temporal information with content information or treat it as a secondary aspect of the recommendation process.

\subsection{Modeling Temporal Order}
\subsubsection{Temporal Embedding}
To effectively encode temporal information in user interaction sequences, we introduce temporal embedding—a cross-token level position embedding approach that preserves the semantic integrity of each item while explicitly representing its position in the user's interaction history.
Specifically, for sequence $\mathcal{S}$, we assign each item $i_k$ a position embedding $p_k$ based on its relative position $k$ (\textit{i.e.}, its distance to the most recent item)in interaction sequence, and add $p_k$ to every token of that item.
Crucially, rather than applying this embedding at the token level (as in traditional Transformer models), we add the same position embedding $p_k$ to every token that constitutes the item. This approach treats each item as a cohesive temporal unit, reinforcing its semantic boundaries within the sequence. We also retain the original token-level positional embeddings to preserve the model's ability to understand contextual relationships within the prompt.
Consequently, the final input embedding for the t-th token is:
\setlength{\abovedisplayskip}{4pt}
\begin{equation}
    \label{eq:inputembed}
I_t=x_t+p_t+p_k,
\end{equation}
\setlength{\abovedisplayskip}{4pt}
where $x_t$ and $p_t$ denote the token embedding and position embedding of the $t$-th token respectively, and $p_k$ represents the temporal embedding of the $k$-th item.
Besides, $p_k$ is set to null if the $t$-th token does not belong to any item.
For example, suppose that the movie \emph{Shawshank Redemption} appears in the third position of $\mathcal{S}$, tokenized into 5 tokens (\textit{i.e.}, `Sh', `aw', `sh', `ank', `Redemption'), then each of these 5 tokens will be augmented by the embedding $p_3$, signifying that they collectively represent an item that appeared third in the user's interaction history.

This temporal embedding approach offers two primary advantages over standard position embeddings: 1) Item-level semantic coherence: By applying the same temporal embedding to all tokens of an item, we enable the model to treat each item as a cohesive unit during representation learning and inference. This alignment between semantic boundaries and temporal encoding helps the model better leverage item-level prior knowledge and understand items as integral units in the user's preference evolution. 2)Explicit temporal encoding: By assigning incrementally increasing temporal embeddings to items based on their relative interaction order ($p_9, \ldots, p_2, p_1, p_0 $), we explicitly encode both absolute and relative temporal relationships, allowing later items to have a greater influence on the prediction. This makes the temporal structure of the interaction sequence more salient to the model, strengthening its ability to detect patterns like recent interest shifts or consistent preferences over time.

\subsubsection{Connection to Existing Methods}
The idea of incorporating temporal embeddings has appeared in several existing methods.  
In terms of implementation, however, temporal embeddings differ from other approaches.  
For example, the Target-aware Temporal Encoding (TTE) in TIN adds temporal information to the ID embedding separately within a Transformer-like architecture.  
This design is essentially similar to conventional positional embeddings.  
In contrast, our temporal embedding is uniformly added to each item at the input layer of the LLM, providing richer semantic information.  

The whole-word embedding in P5~\cite{P5} also applies additional embeddings to all tokens of an item.  
However, since it is applied across all tokens in the input sequence, its contribution to modeling item-level temporal information is diluted.  
Moreover, its encoding order increases sequentially with the token positions, just like standard positional embeddings.  
This means that items appearing later in the sequence undergo greater changes, thereby weakening their overall influence on the final prediction.
\subsection{Counterfactual Inference}

Having established our method for encoding temporal information, we now introduce a counterfactual inference framework to enhance the model's sensitivity to this information. Counterfactual inference allows us to answer questions about what would happen under hypothetical conditions that differ from observed reality.

In our context, we perform counterfactual inference by altering the temporal order variable $T$ to construct a counterfactual world where all temporal information is erased, while preserving all other aspects of the recommendation scenario. This approach enables us to isolate and measure the specific impact of temporal ordering on recommendation outcomes.
The counterfactual inference framework evaluates how descendant variables would change under a different treatment variable while holding other conditions constant\cite{cfinfer,cfinfer1}. In our causal graph, this can be expressed as: `\emph{what $Y$ would be in situation $U = u$ and $I = i$, had $T$ been $t^*$?}'. This counterfactual query examines an alternative course of history where the temporal structure of interactions is altered, while keeping all other factors fixed.

Traditional methods for estimating counterfactual effects require computing and storing posterior information for all background variables in the model—a requirement that is often impractical for complex systems. To address this challenge, we employ the twin network approach, a graphical method that uses two structurally identical networks to jointly represent the factual and counterfactual worlds \cite{twinnetwork}. When performing counterfactual inference on a variable, all arrows entering this variable in the counterfactual world are removed to reflect the intervention, allowing the counterfactual query to be computed using standard probabilistic inference techniques.



In our model, the resulting twin network of our counterfactual inference is shown in Figure~\ref{fig:Causal Anal}(b). The counterfactual intervention effectively places all user interactions at the same—and earliest—time step, maximally removing both relative and absolute order information.
Concretely, we implement this counterfactual world by replacing each item's temporal embedding in the input sequence with the temporal embedding of the first item $i_0$, denoted $p_0$. This intervention creates a scenario where all items appear to have occurred at the same point in time, effectively erasing any sequential patterns or temporal dynamics that might influence recommendation outcomes. The resulting counterfactual world thus preserves the exact set of interacted items and all other latent factors, differing only in its lack of temporal structure.

Based on the twin network framework, the causal effect of temporal order can be computed as the difference between the factual and counterfactual predictions:
\begin{equation}
    \label{eq:6}
CE_T =P(Y\ |\ T = t, I = i, U = u)- P(Y^*\ |\ T = t^*, I = i, U = u),
    \end{equation}
where $t$ denotes the original temporal order, while $t^*$ represents the erased (counterfactual) order.
This causal effect quantifies how much the recommendation outcomes change when temporal information is removed, providing a direct measure of the model's sensitivity to temporal patterns.

\subsection{Model Paradigm}
Building on the causal analysis and counterfactual inference framework described above, we now present the complete CETRec model paradigm for both the tuning phase and inference phase.

\vspace{10pt}
\noindent\textbf{Tuning.}
The core insight of CETRec is that we can enhance a model's temporal sensitivity by explicitly optimizing for the divergence between predictions made with and without temporal information. Based on the causal effect estimated via counterfactual inference, the influence of temporal order on the model's token generation probability can be expressed as:
$$
f_\Theta(x_t, y_{<k}) \;-\; f_\Theta(x_0, y_{<k}),
$$
where $f_\Theta$ represents the model's probability distribution over the next token, $x_t$ denotes the instruction input with temporally ordered sequence,  $x_0$ denotes the same input with the temporally erased sequence (the counterfactual world), and $y_{<k}$ represents the generated tokens up to position $k$.

To encourage the model to become more sensitive to temporal information, we design a novel counterfactual tuning loss $L_{CT}$ that directly optimizes this influence:
\begin{equation}
    \label{eq:7}
L_{CT} = \sum_{(x,y) \in \mathcal{D}} \sum_{k=1}^{|y|} \ell \left( f_{\Theta+\Phi}(x_t, y_{<k}) - f_{\Theta+\Phi}(x_0, y_{<k}); y_k \right).
\end{equation}
This loss function encourages the model to produce different predictions when temporal information is altered, effectively teaching it to recognize and utilize both absolute and relative temporal signals in user interaction sequences. By maximizing the divergence between predictions in the factual and counterfactual worlds, we directly optimize the model's temporal sensitivity.

In addition to the counterfactual tuning loss, we also define a temporal-aware loss $L_{TA}$that enhances the model's overall recommendation capability using the temporally ordered input:
\begin{equation}
\label{eq:8}
L_{TA}=\sum_{(x,y)\in D}\sum_{k=1}^{|y|}\ell\left(f_{\Theta+\Phi}(x, y_{<k});\ y_k\right).
\end{equation}
This loss function ensures that the model maintains high recommendation accuracy while developing increased sensitivity to temporal patterns. By combining these two objectives, we create a balanced training regime that improves temporal awareness without sacrificing overall recommendation quality.

The formal objective of CETRec is thus defined as a weighted combination of these two loss functions:
\begin{equation}
\label{eq:9}
    L= L_{TA} + \lambda L_{CT}
\end{equation}
where $\lambda$ controls the importance of counterfactual loss.
This weighting allows us to adjust the trade-off between optimizing for recommendation accuracy and temporal sensitivity, enabling flexible adaptation to different recommendation scenarios and user preference patterns.

\vspace{10pt}
\noindent\textbf{Inference.}
During the inference phase, the model generates predicted items in natural language format based on the instruction input, using the temporally ordered sequence with original temporal embeddings. This approach ensures that the model leverages its enhanced temporal sensitivity to produce recommendations that account for both the content of previous interactions and their temporal arrangement.
By preserving the temporal structure during inference, CETRec can effectively capture evolving user preferences, recent interest shifts, and time-dependent patterns in interaction sequences. This temporal awareness enables more personalized and contextually relevant recommendations that align with users' current interests rather than outdated preferences.

\subsection{Discussion}
Our methodological design for CETRec highlights several key considerations in temporal modeling for LLM-based recommendation. 
The temporal embedding approach strikes a balance between preserving language understanding capabilities and enhancing temporal awareness by treating items as cohesive units. 
The observed differences between SinPE and RoPE implementations—where RoPE achieves better recommendation performance while SinPE shows higher temporal sensitivity—reflect their distinct mathematical properties: RoPE excels at capturing nuanced relative relationships, while SinPE creates more pronounced absolute position signals. 
Furthermore, our counterfactual inference framework connects to broader trends in causal machine learning, offering advantages over purely correlational methods by explicitly isolating temporal effects. 

\begin{table}[t]
  \centering
  \renewcommand\arraystretch{1.2}
  \caption{Statistical details of training datasets.}
  \label{exp:dataset}
  \begin{tabular}{lccc}
    \toprule
    Dataset     &MovieLens&Steam&LastFM   \\
    \midrule
    \#Sequences    &192,483  &151,056&46,897 \\
    \#Items        &  3,952  &  3,581&4,606 \\
    \#Interactions &999,611  &239,796& 73,510\\
    \bottomrule
  \end{tabular}
\end{table}

\section{Experiment}

In this section, we conduct experiments to answer the following research questions:
\begin{itemize}[leftmargin=*]
  \item \textbf{RQ1}: How does CETRec perform compared to other traditional and LLM-based recommenders?
  \item \textbf{RQ2}: How does different position embedding methods applied in temporal embedding perform?
  \item \textbf{RQ3}: What are the respective contributions of the two components of CETRec to its performance?
  \item \textbf{RQ4}: How effective is CETRec at enhancing an LLM’s sensitivity to temporal information?
\end{itemize}

\begin{table*}[h]
  \centering
  \renewcommand\arraystretch{1.2}
  \setlength{\tabcolsep}{3pt} 
  \small
  \caption{Performance of baselines and CETRec on MovieLens, Steam, and LastFM datasets.}
  \label{tab:performance}
  \begin{tabular}{lcccc|cccc|cccc}
    \toprule
     &\multicolumn{4}{c|}{MovieLens} 
     &\multicolumn{4}{c|}{Steam} 
     &\multicolumn{4}{c}{LastFM} \\
    \cmidrule(lr){2-5} \cmidrule(lr){6-9} \cmidrule(lr){10-13}
     &HR@5&NDCG@5&HR@10&NDCG@10 
     &HR@5&NDCG@5&HR@10&NDCG@10
     &HR@5&NDCG@5&HR@10&NDCG@10 \\
    \midrule
Caser          &0.0667&0.0399&0.1217&0.0573&0.0307&0.0200&0.0488&0.0258&0.0222&0.0145&0.0324&0.0177 \\
GRU4Rec        &0.0475&0.0283&0.0900&0.0417&0.0301&0.0198&0.0479&0.0255&0.0130&0.0079&0.0204&0.0103 \\
SASRec         &0.0542&0.0322&0.0917&0.0441&0.0269&0.0171&0.0450&0.0229&0.0194&0.0097&0.0352&0.0146 \\
P5             &0.0538&0.0334&0.0911&0.0452&0.0688&0.0468&0.1049&0.0583&0.0173&0.0138&0.0273&0.0171 \\
BIGRec
&0.0985&0.0614&0.1093&0.0650&0.0778&0.0530&0.0979&0.0596&0.0264&0.0192&0.0356&0.0221 \\
E4SRec
&0.1050&0.0660&0.1170&0.0700&0.0810&0.0555&0.0982&0.0615&0.0500&0.0325&0.0600&0.0355 \\
LLaRA
&0.1120&0.0710&0.1250&0.0750&0.0850&0.0580&0.0985&0.0635&0.0745&0.0450&0.0880&0.0490 \\
CFT
&0.1043&0.0623&0.1241&0.0691&0.0879&0.0606&0.1013&0.0650&0.0711& 0.0424&0.0838&0.0467 \\

\midrule
CETRec (SinPE) &0.1101&0.0693&0.1209&0.0728&0.0670&0.0436&0.0779&0.0471&0.0780&0.0520&0.0910&0.0550 \\
CETRec (RoPE)  &\textbf{0.1283}&\textbf{0.0779}&\textbf{0.1365}&\textbf{0.0806}&\textbf{0.0921}&\textbf{0.0641}&\textbf{0.1072}&\textbf{0.0690}&\textbf{0.0866}&\textbf{0.0591}&\textbf{0.1030}&\textbf{0.0646} \\
    \bottomrule
  \end{tabular}
\end{table*}

\subsection{Experiment Settings}

\subsubsection{Datasets}
We conduct experiments on three real real-wrold datasets: MovieLens\footnote{\url{https://grouplens.org/datasets/movielens/}} , Steam \cite{SASRec} and LastFM \cite{lastfm}. MovieLens dataset is a well-known movie recommendation dataset containing the movies titles, user ratings and timestamps. Steam dataset is a video-games recommendation dataset that records the title of games users have played on the Steam Store. LastFM dataset contains the name of artists users listen to collected from the Last.fm music platform. In both datasets, user interactions can be organized into sequences in chronological order.

Due to the massive number of parameters in large language models, tuning an LLM consumes substantial time and computational resources, so the dataset size must be kept within a reasonable range. Therefore, for MovieLens dataset, we choose MovieLens1M dataset and sample $1/10$ of interactions. For the Steam dataset, we first retained users who had played at least 20 games, and then randomly sampled $1/3$ of users and $1/3$ of games. For the two datasets, we construct, interaction sequences of length 3 to 10 based on each user's item interactions. Then we sort all sequences in chronological order and then split the data into training, validation, and testing data at the ratio of 8:1:1. This strategy ensures that future interactions are excluded from the training set, thereby preventing information leakage. The statistical details of training datasets are summerized in Table~\ref{exp:dataset}.

\subsubsection{Evaluation Settings}
For each sequence, we designate the item immediately interacted after it as the target. We employ all-ranking protocol for evaluation, treating all non-interacted items as potential candidates and select two widely used metrics: Hit Ratio (HR@K) and Normalized Discounted Cumulative Gain (NDCG@K) \cite{SASRec, bigrec}, with $K \in \{5,10\}$. 
\subsubsection{Baselines}
We demonstrate the ability of CETRec by comparing it to the following traditional and LLM-based methods:

\begin{itemize}[leftmargin=*]
\item \textbf{GRU4Rec} \cite{GRU4Rec}.  
GRU4Rec is an RNN-based recommender that uses Gated Recurrent Units (GRUs) to encode user interaction histories.

\item \textbf{Caser} \cite{Caser}.  
Caser is a CNN-based model that represents the last \(L\) interactions as an \(L \times d\) embedding “image” and applies horizontal and vertical convolutional filters.

\item \textbf{SASRec} \cite{SASRec}.  
SASRec is Transformer-based recommender that employs masked multi-head self-attention to attend to all previous items in a sequence.

  \item \textbf{P5} \cite{P5}.  
    5 unifies multiple recommendation tasks into a natural-language generation paradigm, representing items as textual IDs and casting recommendation as conditional text generation.
  \item \textbf{BIGRec} \cite{bigrec}.  
    BIGRec converts a user’s interaction history sequence into a textual prompt. The model generates a natural‐language continuation, and a grounding module maps the generated text back to concrete item. 
    
    \item \textbf{E4SRec} \cite{e4s}.  
    E4SRec projects item IDs into LLM input embeddings and maps the LLM outputs back to the item space.
    
  \item \textbf{LLaRA} \cite{llara}.  
    LLaRA combines embeddings learned by traditional sequential recommenders and textual expressions of items as prompt to generate the next prediction.

  \item \textbf{CFT} \cite{CFT}.  
    CFT is a method that enhances modeling of user behavior sequences through counterfactual fine-tuning. While CFT improves overall sequence understanding, it does not explicitly encode the temporal ordering of interactions.
\end{itemize}


\subsubsection{Implementation Details}
 For traditional methods, we use one self-attention blocks in SASRec and one GRU layer in GRU4Rec, other settings follow the original papers. For LLM-based models, we select Llama3-8B \cite{llama3} as the backbone and use the AdamW \cite{adamw} optimizer with a learning rate of $1 \times 10^{-4}$. Specifically, we implement Llama-based P5 in OpenP5 paper \cite{openp5, index}. For CETRec, we adopt the two position encoding schemes introduced earlier\textemdash SinPE and RoPE\textemdash as our implementations of temporal embeddings. Besides, we tune the $\lambda$, the hyperparameter controlling the weight of counterfactual loss in our CT task, in the range $\{0.01, 0.1, 0.5, 1, 10\}$. In inference stage, we follow the process in CFT paper \cite{CFT}. We generate five items, finding the most closely matched actual item for each of them to generate the Top-5 recommendation list. The Top-10 recommendation list is generated from the second-best matched items of each item.
All experiments are conducted with Python 3.12, PyTorch 2.2.2 and transformers 4.43.4.

\begin{table*}[h]
  \centering
  \renewcommand\arraystretch{1.2}
    \setlength{\tabcolsep}{3pt}
  \small
  \caption{Ablation results with different positional embedding methods and training variants.}
  \label{tab:abla}
  \begin{tabular}{l lcccc|cccc|cccc}
    \toprule
     &\multicolumn{5}{c|}{MovieLens} 
     &\multicolumn{4}{c|}{Steam} 
     &\multicolumn{4}{c}{LastFM} \\
    \cmidrule(lr){3-6} \cmidrule(lr){7-10} \cmidrule(lr){11-14}
     &
     &HR@5 &NDCG@5&HR@10&NDCG@10
     &HR@5 &NDCG@5&HR@10&NDCG@10
     &HR@5 &NDCG@5&HR@10&NDCG@10 \\
    \midrule
    \multirow{2}{*}{SinPE}
     &CETRec
     &0.1101&0.0693&0.1209&0.0728&0.0670&0.0436&0.0779&0.0471&0.0780&0.0520&0.0910&0.0550 \\
     &\textit{w.o.} CT
     &0.1035&0.0634&0.1167&0.0676&0.0620&0.0407&0.0720&0.0440&0.0721&0.0482&0.0865&0.0514 \\
    \midrule
    \multirow{2}{*}{RoPE}
     &CETRec
     &\textbf{0.1283}&\textbf{0.0779}&\textbf{0.1365}&\textbf{0.0806}
     &\textbf{0.0921}&\textbf{0.0641}&\textbf{0.1072}&\textbf{0.0690}
     &\textbf{0.0866}&\textbf{0.0591}&\textbf{0.1030}&\textbf{0.0646} \\
     &\textit{w.o.} CT
     &0.1192&0.0745&0.1300&0.0781&0.0856&0.0613&0.1021&0.0669&0.0805&0.0565&0.0980&0.0626 \\
    \midrule                
     &\textit{w.o.} Both
     &0.0985&0.0614&0.1093&0.0650&0.0778&0.0530&0.0979&0.0596&0.0264&0.0192&0.0356 &0.0221 \\
    \bottomrule
  \end{tabular}
\end{table*}

\subsection{Performance Comparison(RQ1 \& RQ2)}
We first compare the recommendation performance of baseline methods and CETRec implemented with SinPE and RoPE. The evaluation results are summerized in Table~\ref{tab:performance}. We draw the following observations from the results:
\begin{itemize}[leftmargin=*]
  \item CETRec with RoPE outperforms all the baseline models on three datasets, achiving improvement across all metrics and datasets. This shows the effectiveness of CETRec.
  \item There remains a performance gap between traditional and LLM-based methods. This is partly because traditional models rely on ID embeddings to capture sequential patterns, thereby neglecting the rich semantic information that LLMs leverage for deeper item understanding. Additionally, the orders-of-magnitude difference in parameter scale between traditional models and LLMs contributes to their differing inference capabilities.
  \item P5 underperforms most LLM‑based methods across several metrics, primarily because it represents items by their IDs and thus sacrifices the deeper semantic understanding that LLMs can provide. However, as $K$ increases, P5’s gains outpace those of other LLM‑based models. This is explained by its inference strategy: P5 generates a list of 10 items for a Top‑10 recommendation, rather than 5 items in compared methods, which yields greater recommendation diversity. This behavior is consistent with observations reported in the CFT paper \cite{CFT}.
  \item CETRec implemented with RoPE performances better than with SinPE. This discrepancy can be attributed to two factors. First, by Equation~\ref{eq:rope}, RoPE’s effect depends only on the difference between tokens’ position embedding indices $n$ and $m$. In interaction sequences, items that occur earlier receive temporal embedding indices that are farther from the current position, resulting in larger rotation angles and greater deviation from their original representations. This semantic‑order–based encoding of temporal distance is not captured by SinPE. Second, the Llama architecture \cite{llama3} itself employs RoPE as its positional embedding scheme, enabling a more seamless integration with LLM internals compared to SinPE.
\end{itemize}

\subsection{Ablation Study(RQ3)}

To enhance the temporal sensitivity of LLMs for sequential recommendation, CETRec firstly proposes temporal embedding and accordingly designs the counterfactual tuning task. After demonstrating the model’s efficacy, we conduct ablation experiments to validate the effectiveness of its two core components. We introduce two variants:
\begin{itemize}[leftmargin=*]
  \item \textbf{\textit{w.o.} CT} disables the counterfactual tuning and only allows tuning by the temporal-aware loss $L_{TA}$ in Equation~\ref{eq:8}, this is equivalent to setting $\lambda$ to 0 in Equation~\ref{eq:9}.
  \item \textbf{\textit{w.o.} Both} variant disables both components. Since counterfactual tuning depends on the presence of temporal embeddings, we cannot disable temporal embeddings independently. However, we can recognize the effect of temporal embedding via camparing the two variants. Notably, under our experimental setup, this variant is effectively equivalent to the vanilla BIGRec method.
\end{itemize}
The results are detailed in Table~\ref{tab:abla}, from which we can observe:
\begin{itemize}[leftmargin=*]
  \item Disabling counterfactual tuning (\textit{w.o.} CT) results in a significant performance drop across all datasets and metrics, confirming that this tuning method strengthens the LLM’s understanding of user interaction sequences.
  \item The performance changes of \textit{w.o.} Both variant explains the gap between BIGRec and CETRec. Compared to the results of \textit{w.o.} CT, the differences prove the effectiveness of temporal embedding.
  \item From these performance differences, we conclude that—whether using SinPE or RoPE—both the temporal embedding module and the counterfactual tuning component are critical for CETRec.
\end{itemize}

\begin{figure}[t]
  \centering
  \includegraphics[width=\linewidth]{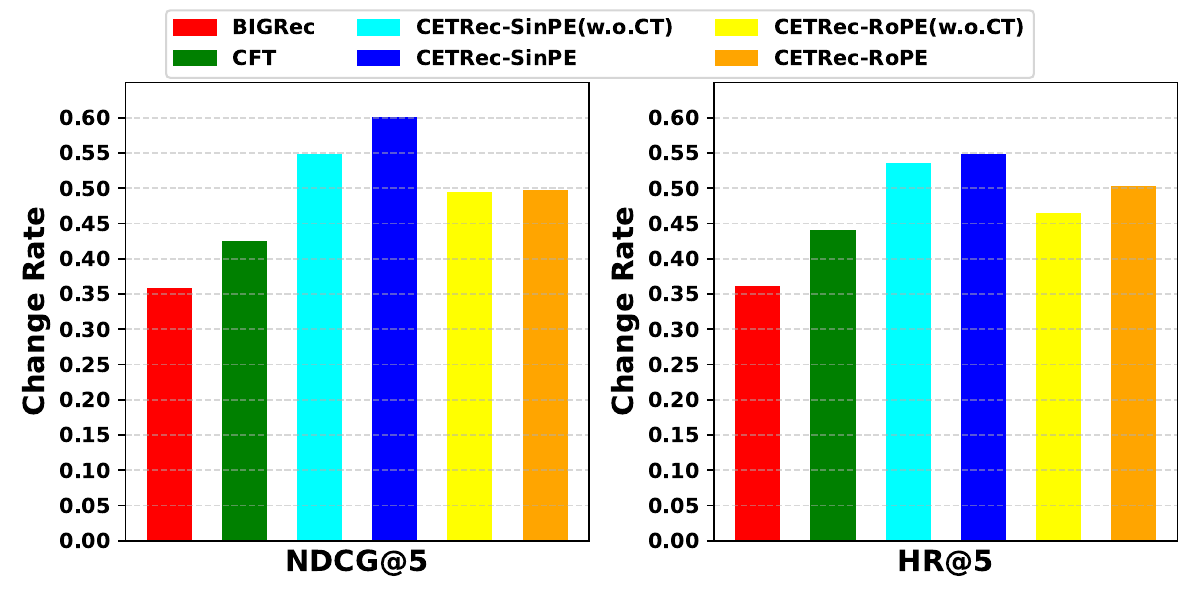}
  \vspace{-10pt}
  \caption{The performance change rate of each methods testing on reversed sequences in test set of MovieLens datasets.}
  \vspace{-10pt}
  \label{fig:5}    
  \Description{Results to illustrate models' temporal sensitivity.}
\end{figure}

\subsection{Temporal Sensitivity Analysis(RQ4)}
We further conduct experiments to investigate the temporal sensitivity of CETRec. By reversing the interaction sequences, we alter both the absolute order and the relative order of each sequence, thereby changing the temporal information they contain. Therefore, we evaluate the model’s performance on the reversed interaction sequences in the test set and quantify its temporal sensitivity by measuring the change rate to its performance on the original test set. BIGRec and CFT are considered as baselines because of their similar paradigm, effectively demonstrating the impact of different methods on temporal sensitivity of models. We use the MovieLens dataset as an example. The results are shown in Figure~\ref{fig:5}, and we draw the following conclusions:
\begin{itemize}[leftmargin=*]
  \item Both CETRec and the \textit{w.o.} CT variant increase the performance change rate, demonstrating the effectiveness of both counterfactual tuning and temporal embeddings. However, CETRec’s additional gain over the \textit{w.o.} CT variant is smaller. This is partly because counterfactual tuning builds on the sequence order captured by temporal embeddings, and partly because, in a recommendation sequence composed of both items and temporal signals, the latter contribute only so much—so the change rate cannot grow without bound.
  \item CFT’s change rate is larger than BIGRec’s but still lower than CETRec’s. This is because CFT improves the model’s understanding of the overall interaction sequence, but its temporal modeling granularity is not as fine‐grained as that of CETRec.
  \item CETRec with SinPE exhibits a higher change rate than with RoPE. This is because SinPE, as an absolute position encoding, induces larger and more consistent shifts in token embeddings, causing the model to react more strongly to differences in temporal embeddings and thus demonstrate greater temporal sensitivity. This also shows that higher temporal sensitivity does not necessarily translate into better recommendation performance.
\end{itemize}

\subsection{Case Sduty}

\begin{figure}[t]
  \centering
  \includegraphics[width=\linewidth]{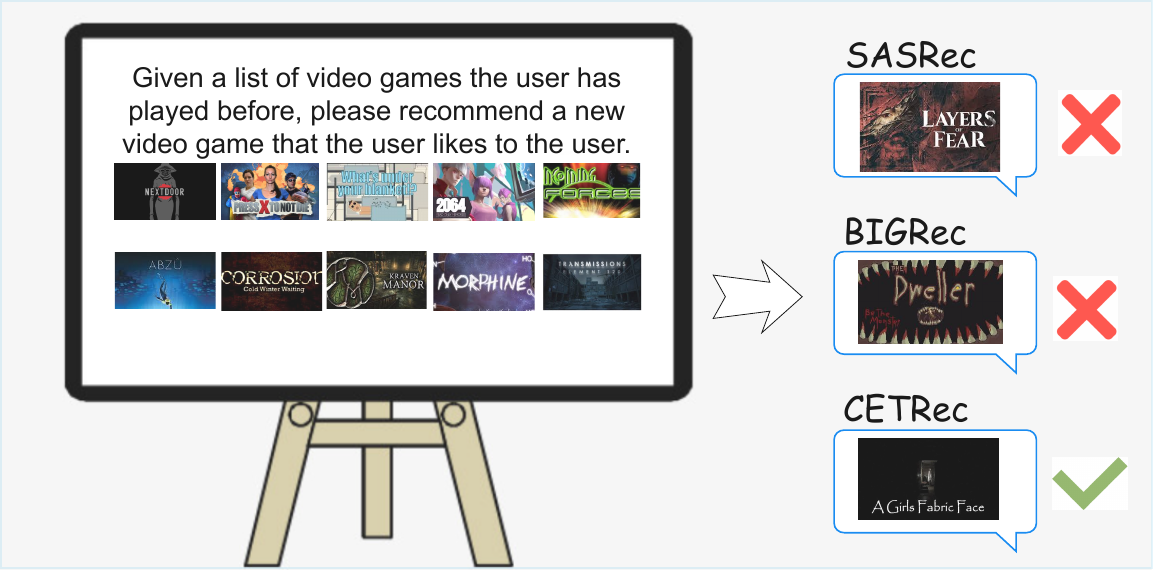}
  \vspace{-10pt}
  \caption{Case study on Steam Dataset.}
  \vspace{-10pt}
  \label{fig:case}    
  \Description{Case study on Steam Dataset.}
\end{figure}

To illustrate the effectiveness of CETRec and its ability to exploit temporal information in interaction sequences, we conduct a case study on the Steam dataset, comparing SASRec, BIGRec, and CETRec with RoPE.  
As shown in Figure~\ref{fig:case}, for a user who sequentially played 
\textit{"The Next Door"}, \textit{"Press X to Not Die"}, \textit{"What's Under Your Blanket!?"}, 
\textit{"2064: Read Only Memories"}, \textit{"Incoming Forces"}, \textit{"ABZU"}, 
\textit{"Corrosion: Cold Winter Waiting [Enhanced Edition]"}, \textit{"Kraven Manor"}, 
\textit{"Morphine"}, and \textit{"Transmissions: Element 120"}:  
SASRec recommended "Layers of Fear", following the genres of psychological horror adventure games.  
BIGRec considered the content of the above games and recommended \textit{"The Dweller"}.  
In contrast, CETRec captured the user's evolving interest in both horror adventure and puzzle games over time, and correctly recommended \textit{"A Girls Fabric Face"}.
\section{Conclusion}

In this work, we identify the limited temporal awareness of LLM-based sequential recommenders and propose CETRec, a causal inference–driven framework that treats temporal order as an independent causal factor. By applying item-level temporal embeddings and counterfactual tuning, CETRec significantly improves temporal sensitivity and achieves consistent gains over mutipule baselines on real-world datasets. Future work could extend this framework by incorporating richer temporal signals (e.g., time intervals, seasonal patterns) and applying counterfactual reasoning to other recommendation factors such as user context or item popularity.

\newpage

\bibliographystyle{ACM-Reference-Format}
\bibliography{reference}

\end{document}